\documentclass{article}
\usepackage{siunitx}
\usepackage{booktabs}
\usepackage{multirow}
\usepackage[table]{xcolor}
\usepackage[preprint]{corl_2026} %

\usepackage{moreverb,url}
\usepackage{xspace}
\usepackage{siunitx}
\usepackage{wrapfig}

\usepackage{amssymb}
\usepackage{amsmath}
\usepackage{amsthm}
\usepackage{amsbsy}
\usepackage{bbm}
\usepackage{dsfont}

\usepackage{amsmath}
\usepackage{amssymb}
\usepackage{graphicx}
\usepackage{algorithm}
\usepackage{algpseudocode} 

\usepackage{graphicx}
\usepackage{wrapfig}
\usepackage{booktabs} %

\usepackage{subfiles}
\usepackage{csquotes}
\usepackage{bm}

\usepackage{tocloft}
\usepackage{array}

\usepackage{graphicx}
\usepackage{subfig} 
\usepackage{multirow}

\newcommand\BibTeX{{\rmfamily B\kern-.05em \textsc{i\kern-.025em b}\kern-.08em
T\kern-.1667em\lower.7ex\hbox{E}\kern-.125emX}}

\definecolor{wine}{RGB}{204, 0, 102}
\definecolor{magenta_wine}{RGB}{158, 44, 143}
\definecolor{dusty_wine}{RGB}{143, 59, 101}
\definecolor{ocean}{RGB}{13, 121, 202}
\definecolor{light_ocean}{RGB}{18, 178, 235}
\definecolor{dark_ocean}{RGB}{10, 89, 148}
\definecolor{grey}{RGB}{170, 170, 170}
\definecolor{light-grey}{RGB}{220, 220, 220}
\definecolor{dark-gray}{rgb}{0.2, 0.2, 0.2} 
\definecolor{med-grey}{rgb}{0.3, 0.3, 0.3} 
\definecolor{grape}{RGB}{112,48,160}
\definecolor{aqua}{RGB}{52,172,139}
\definecolor{dark_aqua}{RGB}{35,115,93}
\definecolor{dark_orange}{RGB}{216,92,0}
\definecolor{vibrant_orange}{RGB}{250, 160, 26}
\definecolor{vibrant_blue}{RGB}{14, 120, 255}
\definecolor{vibrant_pink}{RGB}{255, 0, 104}
\definecolor{dark_red}{RGB}{122, 0, 0}
\definecolor{dark_green}{RGB}{0, 92, 34}
\definecolor{dusty_blue}{RGB}{77, 91, 128}
\definecolor{dark_brown}{RGB}{125, 54, 36}
\definecolor{violet}{RGB}{116, 12, 173}

\definecolor{green-reg}{RGB}{13, 148, 143}
\definecolor{blue-reg}{RGB}{49, 131, 189}
\hypersetup{
    colorlinks=true,
    linkcolor=vibrant_orange, %
    citecolor=vibrant_orange, %
    urlcolor=vibrant_orange   %
}

\newcommand{\ours}{\textcolor{vibrant_orange}{\texttt{ELASTIC}}\xspace}
\newcommand{\fixed}{\textcolor{med-grey}{\textbf{Fixed}}\xspace}
\newcommand{\seq}{\textcolor{dark_aqua}{\textbf{Sequential}}\xspace}
\newcommand{\dip}{\textbf{Diffusion Policy}\xspace}
\newcommand{\bon}{\textcolor{blue-reg}{\textbf{BoN}}\xspace}

\newcommand{\ntime}{\tau} %
\newcommand{\state}{s} 
\newcommand{\stateSpace}{\mathcal{S}}
\newcommand{\action}{a}
\newcommand{\actionSpace}{\mathcal{A}}
\newcommand{\stepsize}{\Delta \tau}
\newcommand{\dtime}{k}
\newcommand{\ptime}{t} %
\newcommand{\pstate}{x}
\newcommand{\pstateSpace}{\mathcal{X}}
\newcommand{\ctrl}{u} %
 
\newcommand{\ctrlSpace}{\mathcal{U}}

\newcommand{\rew}{r}
\newcommand{\rewtask}{r^\texttt{task}}
\newcommand{\rewcomp}{r^\texttt{compute}}
\newcommand{\qfun}{Q^\texttt{task}}

\newcommand{\metamdp}{\mathcal{M}^{\texttt{meta}}}
\newcommand{\mask}{m}

\newcommand{\mpolicy}{\pi^\texttt{{meta}}} %
\newcommand{\dpolicy}{\pi^\texttt{{GCP}}} %

\newcommand{\mcritic}{Q^\texttt{meta}}

\title{\ours: Efficiently Learning to Adaptively Scale Test-Time Compute for Generative Control Policies}

\author{
  Andrew Zou Li, Gokul Swamy, Yonatan Bisk, Andrea Bajcsy \\
  Carnegie Mellon University \\
  \texttt{\{azl2, gswamy, ybisk, abjacsy\}@andrew.cmu.edu}
}

\usepackage[skip=5pt]{caption} %
\usepackage{wrapfig}

\begin{document}
\maketitle

\begin{abstract}
Generative control policies (GCPs), such as diffusion policies and flow-based vision-language-action models, enable test-time scaling in robot control. Test-time compute can be allocated along two axes: \textit{sequential scaling}, which increases denoising steps to refine actions, and \textit{parallel scaling}, which samples multiple candidate actions to search across modes of the policy distribution. 
However, the optimal allocation of sequential and parallel compute is hard to know a priori as it is state-, task-, and policy-dependent. For example, early stages of a grasp may benefit from broader parallel exploration, while near-contact phases may require more sequential refinement for precision. 
We present \ours, an algorithm that learns state-dependent test-time \textit{compute schedules} for GCPs. We formulate compute allocation as a meta-Markov Decision Process in which a meta-policy interacts with a frozen pretrained robot policy and selects sequential steps and parallel samples at each denoising iteration to maximize task success while minimizing compute. 
Using reinforcement learning, this meta-policy also learns adaptive compute schedules without access to the GCP's training data. Across simulated manipulation benchmarks with diffusion policies, \ours Pareto-dominates fixed and single-axis scaling baselines at matched compute budgets. On real-world robot manipulation with the $\pi_{0.5}$ vision-language-action model, \ours matches best-of-$10$ success while reducing wall-clock latency by 34\%.
\end{abstract}
 
\section{Introduction}  

Generative control policies (GCPs), such as diffusion policies and flow-based Vision-Language Action models (VLAs), have demonstrated remarkable success at imitating complex, multimodal distributions over actions \cite{chi2023diffusionpolicy,   pmlr-v305-black25a, pmlr-v270-kim25c, octo_2023}. At inference time, these models iteratively transform samples from a Gaussian distribution to the target action distribution via a sequential refinement process \cite{Ho_Jain_Abbeel_2020, Song_Meng_Ermon_2020, lipman_flow_2023}. This structure is precisely what enables \textbf{test-time scaling} for GCPs: the ability to improve policy performance via more inference-time computation, rather than via more training.

\begin{figure}
    \centering
    \includegraphics[width=0.9\linewidth]{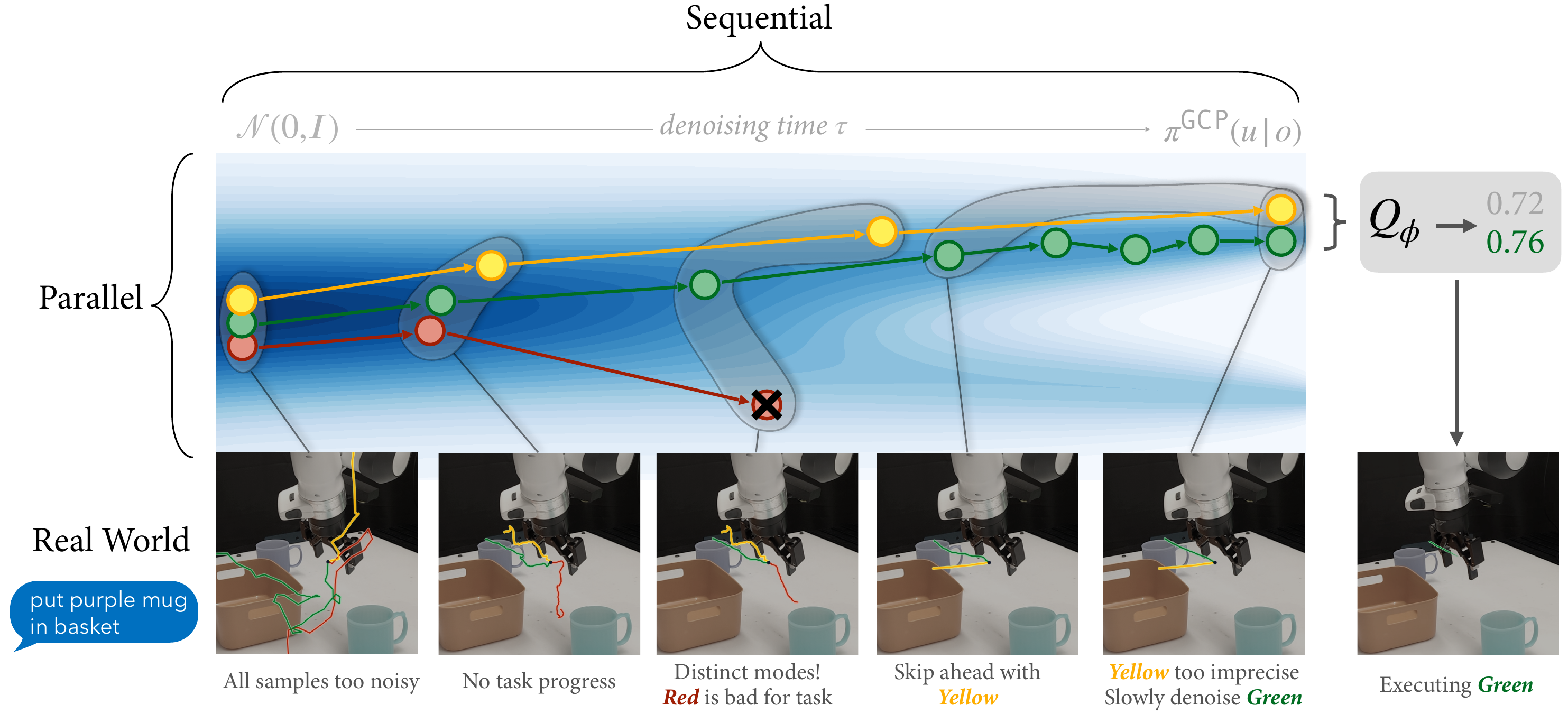}
    \caption{\textbf{Parallel vs. Sequential Scaling in Robotics.} Parallel scaling explores diverse action modes, while sequential scaling refines solution quality. Our \ours~algorithm jointly allocates both compute axes to optimize the performance--latency tradeoff.}
    \vspace{-1em}
    \label{fig:front-fig}
\end{figure}

Broadly speaking, test-time compute can be allocated along two orthogonal axes.
\textbf{Sequential scaling} refines a robot's action samples toward higher-quality solutions through additional integration steps \cite{Song_Meng_Ermon_2020}. For example, consider the pick-and-place manipulation task in Figure~\ref{fig:front-fig}. Finer integration steps during denoising yield more precise robot motions. 
\textbf{Parallel scaling} samples and denoises multiple candidates from the policy, exploring different modes of the action distribution. 
From these, one can select the best sample for a given task via a verifier (e.g., Best-of-$N$ search \cite{brown2025large}). 
Going back to  Figure~\ref{fig:front-fig}, parallel scaling enables the robot to explore action modes that move towards the purple or the blue mug; the verifier then discards the blue mug action mode based on the prompt.

Yet, it is not obvious a priori which axis to scale: the right allocation varies with the task, the state, and the policy's own competence. 
For example, sequential scaling is ineffective on the mug pick-and-place task when the base policy is uncertain about which mug to pick up; this is fundamentally a mode selection problem that benefits more from parallel sampling. 
Conversely, parallel scaling without sufficient sequential refinement produces imprecise actions that can lead to grasp failures. %

One might hope to maximally scale both sequential and parallel compute to maximize performance. 
In robotics, however, test-time computation directly translates into control latency, which is undesirable for real-time, reactive control.
Sequential scaling clearly increases latency by requiring additional denoising steps before action execution. But parallel scaling is also not free: while candidates can be evaluated concurrently in principle, onboard deployment is limited by finite hardware parallelism and memory bandwidth. 
As a result, both test-time scaling axes incur nontrivial latency costs, yet existing methods typically fix compute at conservative and inefficient settings.

Our key insight is that \textit{the optimal test-time compute allocation can be learned via reinforcement learning: by interacting with the frozen base policy, a ``meta-policy'' can learn when to favor sequential scaling or parallel scaling as a function of the task, state, and the GCP’s capabilities.} Specifically, we formulate compute allocation as a meta-Markov Decision Process (MDP), before detailing the key algorithmic choices required to solve this meta-MDP efficiently. We call our algorithm \ours: \textbf{E}fficiently \textbf{L}earning to \textbf{A}daptively \textbf{S}cale \textbf{T}est-t\textbf{i}me \textbf{C}ompute for GCPs.

\noindent \textbf{Contributions.} We formulate compute allocation as a meta-MDP, where a meta-policy interacts with a frozen pretrained GCP and selects the number of sequential steps and parallel samples at each denoising iteration to maximize task success under a compute budget.
In simulated manipulation benchmarks with diffusion policies, \ours Pareto-dominates fixed and single-axis baselines at matched compute budgets. On hardware with the $\pi_{0.5}$ vision-language-action (VLA) model, \ours matches Best-of-$10$ success while reducing wall-clock latency by 34\%.

\section{Related Work}
\label{sec:related-work}
\textbf{Sequential Scaling.} 
Sequential scaling in generative models trades off inference compute and output quality by choosing when to allocate generation steps (e.g., longer reasoning traces in language models or more denoising steps in diffusion models). 
In robotics, this idea has recently been explored for diffusion policies, where inference speed and action quality can be controlled through the number of denoising steps. Prior work includes distillation-based methods \cite{prasad2024consistency} (which compress inference into one or few steps, but require retraining), advanced ODE solvers \cite{Song_Meng_Ermon_2020, lu2022dpm}, and, most relevant to our work, methods that freeze the base policy but instead adapt denoising schedules based on robot state using variance estimation or reinforcement learning \cite{Hu_Liu_Liu_Liu, trivedi2026adaptivetimestepflow, Yu_Gao_Wu_Yu_Wang_2025}. 
We instead study the joint allocation of test-time compute across both sequential and parallel scaling axes, arguing that sequential scaling alone is insufficient for maximizing performance.

\textbf{Parallel Scaling.} An orthogonal axis of test-time scaling is through parallel sampling and verification, as in Best-of-N search \cite{setlur2025scaling, Ma_Tong_Jia_Hu_Su_Zhang_Yang_Li_Jaakkola_Jia_etal._2025, Damani_Shenfeld_Peng_Bobu_Andreas_2024, Qi_Ye_Tang_Zhu_Choi_2025}. 
In robotics, learning to search with verifiers has been explored for improving performance \cite{Kwok_Agia_Sinha_Foutter_Li_Stoica_Mirhoseini_Pavone_2025, jain_smooth_2025, nakamoto2024steering} and aligning policies to human preferences \cite{Wu_Tian_Swamy_Bajcsy_2025, wu2025you}. 
These methods completely denoise a fixed number of candidates, applying a constant parallel overhead regardless of environment state. The resulting overhead is substantial: RoboMonkey \cite{Kwok_Agia_Sinha_Foutter_Li_Stoica_Mirhoseini_Pavone_2025} 
reports over 4$\times$ the base policy inference latency at 16 samples even 
with a specialized serving engine. 
In contrast, we explore interventions during the denoising process of each parallel sample to reduce computational cost while preserving the performance benefits of these methods. 
FASTER~\cite{dong2026fastervalueguidedsamplingfast} attempts to reduce parallel overhead via early stopping, but their proposed noise-aware verifier is in practice used only to select among initial noise samples prior to denoising, rather than to prune candidates mid-trajectory, and does not account for sequential denoising length. 

\textbf{Sequential-Parallel Tradeoff}. Adaptively scaling both sequential and parallel axes is a less studied problem \cite{Qi_Ye_Tang_Zhu_Choi_2025, Manvi_Hong_Seyde_Labonne_Lechner_Levine_2025}.
Most related to our work is the algorithm ZIP-RC \cite{Manvi_Hong_Seyde_Labonne_Lechner_Levine_2025} applied to language modeling, which formulates compute allocation over parallel samples as a meta-MDP, learning to predict the joint distribution over expected reward and the remaining generation length for each sample. 
However, ZIP-RC relies on language models' next-token logits for self-verification, which has no analog in GCPs: actions carry real-world consequences beyond appending a token, and intermediate denoising updates do not encode downstream task success.

\section{Preliminaries: Generative Control Policies \& Test-time Scaling Axes}
\label{sec:prelim}

\noindent \textbf{Diffusion and Flow Matching.} Diffusion and flow-matching models learn to generate samples from a complex target distribution $p(x)$ by transforming samples from an easy distribution.
Specifically, these models learn a time-dependent vector field that, when integrated, transports samples from a Gaussian noise distribution, $\mathcal{N}(0,I)$, to samples from the training distribution, $p(x)$ \cite{Song_Sohl-Dickstein_Kingma_Kumar_Ermon_Poole_2020, Ho_Jain_Abbeel_2020, lipman_flow_2023}.
Let $\theta$ be the parameters of the learned vector field, $\ntime \in [0,1]$ be the denoising time, and $x_\ntime$ be the sample at any denoising time $\ntime$. 
Starting from a noisy sample  $x_1 \sim \mathcal{N}(0,I)$ at $\ntime = 1$, the numerical integration of the learned velocity network $v_\theta$ or noise prediction network $\epsilon_\theta$ results in the denoised sample $x_0 \sim p(x)$ at $\ntime = 0$. In particular, we iteratively apply $x_{\ntime-\stepsize} = f(x_\ntime, \ntime, \stepsize; \theta)$,
where $\stepsize$ is the integration step size.

\noindent \textbf{Generative Control Policies.} We refer to policies that produce actions via an iterative generation process as \textit{generative control policies} (GCPs).
In robotics, diffusion and flow models have been used extensively to model conditional action distributions~\cite{chi2023diffusionpolicy, pmlr-v305-black25a}. 
We denote the robot's policy as 
$\dpolicy := p(\ctrl_{\ptime} \mid o_\ptime)$, where $\ctrl_{\ptime} \in \ctrlSpace$ is a chunk of robot actions (e.g., joint velocities) and $o_\ptime \in \mathcal{O}$ is the robot's current observation (e.g., proprioceptive state and RGB camera images) at time step $\ptime$. 

\noindent \textbf{Scaling Axis 1: Sequential.}
For a fixed denoising step size $\stepsize$, the iterative denoising process requires 
$L = \lceil T / \stepsize \rceil$ evaluations of the learned network ($v_\theta$ or $\epsilon_\theta$). 
Smaller $\stepsize$'s result in finer numerical integration and produce generations that are closer to the target distribution; however, they increase the denoising trajectory length $L$ and thus wall-clock latency.
Larger $\stepsize$'s decrease wall-clock latency, but potentially result in lower-quality generations. 
\textit{\textbf{Sequential test-time scaling involves adapting the denoising step size $\stepsize$ to tradeoff generation fidelity vs. latency.}}

\noindent \textbf{Scaling Axis 2: Parallel.}
Another GCP control knob is the number of action samples that are denoised. 
Intuitively, denoising $N$ independent samples in parallel allows us to hit multiple modes of the learned action distribution. This is commonly coupled with a Best-of-N step, in which each denoised action is evaluated by an external verifier $Q_\phi$ and only the top-performing one is executed. 
However, the use of a verifier and parallel inference increases wall-clock latency under an inference-time memory budget, especially on top of models like VLAs \cite{Kwok_Agia_Sinha_Foutter_Li_Stoica_Mirhoseini_Pavone_2025, Wu_Tian_Swamy_Bajcsy_2025}. 
\textit{\textbf{Parallel test-time scaling adapts the number of samples $N$ that are denoised to tradeoff generation fidelity vs. latency.}}

\section{\ours: Efficiently Learning to Adaptively Scale Test-Time Compute}

Although GCPs expose sequential and parallel test-time compute axes, it is hard to know a priori \textit{which} to scale and \textit{where} to scale. This is because the optimal compute allocation depends on the task (e.g., precise peg insertion requiring greater sequential scaling), the base policy's action distribution (e.g., mode collapse making parallel scaling ineffective), as well as the state the robot is currently in (e.g., sequential scaling being beneficial at critical pre-grasp positions). Thus, we frame the problem of \textit{learning how to allocate test-time compute} as solving a \textit{meta}-Markov Decision Process (MDP).   

\subsection{Test-Time Compute as a \textit{Meta}-Markov Decision Process}
\label{sec:mdp}

We formalize optimizing test-time compute as solving a meta-MDP
$\metamdp = (\stateSpace, \actionSpace, T, \rew)$. We control the joint denoising process over $N$ samples, each corresponding to a denoising trajectory from noise to robot action \cite{dppo2024}. \footnote{While this formulation may make it seem like we're optimizing greedily over a single real-world timestep, our use of a terminal value estimate allows us to optimize over the full real-world episode, as we detail below.} We select one of these actions to execute. 

\noindent \textbf{Physical Environment}. Although our meta-MDP optimizes over the denoising process, the actions generated affect the physical world. Let the physical states be $\pstate \in \pstateSpace$ (e.g., proprioceptive state, environment geometry, object poses) and the robot’s control action (e.g., joint torques) be $\ctrl \in \ctrlSpace$.

\noindent \textbf{States ($\stateSpace$).} Let meta-states $\state_\dtime\in \stateSpace$ be defined as the concatenation of the physical state and denoising state at every denoising step $k$ of the joint denoising trajectory. In particular, we set $\state_\dtime = (\pstate, \ctrl_\dtime^{1:N}, \ntime_\dtime^{1:N}, \mask^{1:N}_\dtime)$ where $\pstate \in \pstateSpace$ is the physical state (constant throughout the denoising process), $\ctrl^{1:N}_\dtime$ are $N$ partially-denoised action samples from the GCP at denoising times $\ntime_\dtime^{1:N}$, and $\mask^{1:N}_\dtime$ is a mask indicating which samples are active. Note that samples may advance to different denoising times $\ntime$ at the same step $\dtime$, so we include the vector $\ntime^{1:N}_\dtime$ as part of the state.

\noindent \textbf{Actions $(\actionSpace)$.} We define meta-actions as $\action := \stepsize_\dtime^{1:N} \in \actionSpace$, which represents how much each sample advances (i.e., the stride of the integration). 
For each sample $i \in |N|$, the action lies in $\stepsize_\dtime^{(i)} \in [0, \ntime_\dtime]$, where we define $\stepsize_\dtime^{(i)} = 0$ as permanently stopping the sample, akin to pruning or early stopping the sample from the search. While the maximum $N$ is fixed, we can scale down parallel compute by removing or never using samples in the parallel search.
In contrast to prior works~\citep{Yu_Gao_Wu_Yu_Wang_2025, dong2026fastervalueguidedsamplingfast}, our choice of action space allows us to \textit{simultaneously optimize} over both axes of test-time scaling in a state-dependent fashion: we control how quickly samples ``jump'' through the denoising process (sequential scaling), as well as keeping/pruning samples (parallel scaling).

\noindent \textbf{Transition $(T)$.} The transition function $T(\state' \mid \state, \action)$ encapsulates both physical and denoising dynamics. Specifically, active action samples $\ctrl_\dtime^{1:N}$ evolve via the denoising dynamics governed by the pretrained $\dpolicy$ with strides $a$. Pruning a sample with $a^{(i)}=0$ sets $m_\dtime^{(i)}=0$. Pruned and fully denoised samples maintain their state $s$ regardless of $a$.
After every sample is fully denoised (i.e., $\ntime_\dtime^{(i)}=0, i \in [N]$) or pruned (i.e., $m^{(i)}_\dtime=0 , i \in [N]$), we feed un-pruned samples to verifier $Q_\phi$. The action with the highest verifier score is executed on the robot, evolving the physical state $\pstate$.

\subsection{Reward Design for Trading Off Performance and Compute}

Our reward is a sum of task performance and computational cost terms. Because nuanced behaviors such as pruning and coordination unfold over the course of the denoising process rather than at a single step, we define $\rewcomp$ as a function of the entire denoising trajectory $\xi = (s_0, a_0, s_1, \dots)$ that culminates in a low-level robot action $u$. 

\noindent \textbf{Task Performance Reward.} Because the denoising process is nested inside the real-world dynamics, the effective horizon of the meta-MDP is quite large. We also only receive a single bit of feedback at the end of the real-world episode. Put together, we are faced with a challenging credit assignment problem. To address this concern, we fit an environment-level $Q$-function for $\rewtask$ and use it as a terminal value estimate at the end of the denoising trajectory. Note that this transformation preserves policy optimality  \cite{ng1999policy}; because $\qfun$ estimates the expected return over the remainder of the real-world episode rather than just the immediate-step reward, it folds the downstream consequences of the current action into a single terminal value. Consequently, optimizing the meta-MDP independently at each robot timestep remains optimal.
We also use this verifier for action selection in parallel scaling (i.e., $Q_\phi=\qfun$). 

\noindent \textbf{Compute Reward.}
We design $\rewcomp$ to be able to flexibly penalize expenditures along either test-time compute axis. First, we define the length of a denoising trajectory as $\ell^{(i)}(\xi) = \sum_\dtime m_k^{(i)}$. Then, our \textit{sequential cost} $L$ is the length of the slowest-finishing sample: $L(\xi) = \max_{i \in [N]}\, \ell^{(i)}$, and our \textit{parallel cost} $P$ is the time-averaged number of active samples: $P(\xi) = \sum_i^N \ell^{(i)}(\xi)/ L(\xi) \geq 1$.
We then scale these costs by weights $\alpha, \beta \in \mathbb{R}_{\geq 0}$ and add the task performance term to arrive at:
\begin{equation}
    r(\xi)
    = \qfun(\pstate,\ctrl)
    - \alpha L(\xi)
    - \beta \left(P(\xi)-1 \right).
    \label{eq:reward}
\end{equation}
The coefficients $\alpha$ and $\beta$ allow us to interpolate between regimes: large $\alpha$ with small $\beta$ favors fast single-sample inference; large $\beta$ with small $\alpha$ penalizes stragglers without penalizing uniform depth; and the magnitudes of both determine the tradeoff between reward and computational cost.  

\subsection{Key Design Decisions for Efficiently Solving the Meta-MDP}
We solve the meta-MDP described in \ref{sec:mdp} via Soft Actor-Critic (SAC) \cite{pmlr-v80-haarnoja18b}, training both a \textit{meta-policy} $\mpolicy: \stateSpace \rightarrow \Delta(\actionSpace)$ and a \textit{meta-critic} $\mcritic: \stateSpace \times \actionSpace \rightarrow \mathbb{R}$. 
The actor is inserted into the base policy's denoising loop and outputs a denoising stride for all $N$ samples following Algorithm~\ref{alg:inference}, while a centralized critic estimates the value of the denoising trajectory under the group reward in Eq.~\eqref{eq:reward}. We now detail several design decisions required to facilitate efficient meta-policy learning. 

First, recall that the meta-MDP state contains a set of $N$ denoising samples whose ordering is arbitrary. Permuting the samples should permute the actor's denoising decisions (equivariance), but should not change the critic's scalar value estimate (invariance). Any architecture that assigns meaning to sample index, such as concatenation, would generalize poorly to identical denoising states appearing in different slots. Thus, we parameterize $\mpolicy$ and $\mcritic$ with attention over the sample set, yielding a permutation-equivariant actor and a permutation-invariant centralized critic.  Second, the samples are coupled by the parallel-compute penalty in Eq.~\eqref{eq:reward}, so the value of denoising one sample depends on other samples. A decentralized design with independent, per sample actors would produce near-symmetric decisions early in denoising. This would prevent desirable coordinated behavior, like a sample in a more promising region of the action space suppressing compute on weaker peers. In response, we use a centralized actor and critic to coordinate strides across samples. \textit{Third}, to ensure strides depend on physical state, we apply FiLM conditioning \cite{perez2018film} to the frozen base policy's observation embedding, modulating how each denoising state is interpreted.

Since solving the meta-MDP is a challenging exploration problem, we improve sample efficiency of training by adopting hybrid reinforcement learning \cite{song2023hybrid}. 
First, in \textbf{Phase 1: Offline Critic Pretraining}, we pretrain the critic on an offline buffer collected with rollouts of $\pi^{\texttt{base}}$ under various fixed denoising schedules. Next, in \textbf{Phase 2: Hybrid RL}, we run rollouts of $\pi^{\texttt{base}}$ with $\pi^{\texttt{meta}}$ controlling the sequential and parallel inference compute following Algorithm~\ref{alg:inference}.
To improve sample efficiency, we augment the replay buffer with counterfactual compute allocations. In particular, we generate this off-policy data by running $\dpolicy$ with all $N$ samples at each state and subsample candidates to simulate varying budgets -- see Appendix~\ref{ap:subsample} for more details.

\section{\ours Outperforms Fixed and Single-Axis Test-Time Scaling}
\label{sec:experiments}

We design our experiments to demonstrate the following: \textbf{(1)} adaptive scaling outperforms fixed allocation across a variety of compute costs and tasks; \textbf{(2)} \ours learns qualitatively reasonable compute allocation behaviors; \textbf{(3)} \ours scales to the state-of-the-art pre-trained $\pi_{0.5}$ VLA on real hardware and improves performance with less wall-clock inference latency.

\subsection{\ours's Test-time Compute Beats Fixed Allocation with Diffusion Policies}
\label{sec:dp}
\noindent \textbf{Setup.} We first demonstrate adaptive allocation outperforms fixed and single-axis scaling under matched compute at three different compute cost combinations. We evaluate on a suite of tasks chosen to surface different mechanisms of train-test mismatch because it is where test-time scaling has clear room to improve performance. We also explore how much we want to scale along each axis as we vary the costs $(\alpha, \beta)$ of scaling each axis in the reward Eq.~\eqref{eq:reward}.

\noindent\textit{Invalidated action modes}: we use \textsc{PushT} \cite{chi2023diffusionpolicy} and a variant called \textsc{PushT Obstacle} designed to invalidate action modes of the base policy by introducing an obstacle at test-time that is not visible to the policy, but is visible in the training data of the verifier.

\noindent\textit{Suboptimal demonstrations}: From Robomimic \cite{robomimic2021}, we use both the \textsc{Square PH} and \textsc{MH} tasks, in which the base policy is trained on higher quality proficient human (PH) data and mixed human (MH) data. 

\noindent\textit{Mode ambiguity}: From Robomimic, we use the \textsc{Can Paired} task, in which the training data has paired 50\% ``good'' (i.e., placing towards the goal location) and 50\% ``bad'' (i.e., dropping the can over the side of the table) mode data. The mode split occurs right after pickup. We create a harder version of the task called \textsc{Can Reverse} by reversing 70\% of the training trajectories at random timesteps after pickup, ensuring $\dpolicy$ is more likely to select a bad mode.

For $\dpolicy$, we use the original Diffusion Policy \cite{chi2023diffusionpolicy} checkpoint when available (\textsc{PushT} and \textsc{Square} tasks) or train a diffusion policy with the default configuration otherwise (\textsc{Can Paired} and \textsc{Can Reverse}). For parallel scaling and the task reward $\qfun$, we train a verifier $Q_\phi$ via Bellman Residual Minimization on a mixture of success and failure data from $\dpolicy$ rollouts (Appendix~\ref{ap:dp-verifier}).

\noindent \textbf{Methods \& Metrics.} We measure the success rate of each method across 200 fixed initial conditions per task. We record the average sequential steps $L$ and average parallel samples $P$ used per action generation. We train and test \ours under three $(\alpha, \beta)$ cost regimes: $(0.03, 0.1)$, $(0.1, 0.03)$, and $(0.1, 0.1)$, covering scenarios where parallel overhead (e.g., memory, verifier cost) or sequential evaluations are the bottleneck, and where both matter.

\noindent \textbf{Baselines.} We compare against baselines that match \ours's average compute $\alpha L + \beta P$, but differ in how that compute is allocated. 
\fixed uses a state-\textit{independent} allocation of $(L, P)$ matched to \ours's mean values, testing whether state adaptation matters beyond choosing the right average allocation. \seq exhausts the budget with only sequential steps on a single sample, testing whether joint allocation matters in each task. For reference, we also report the performance of the default \dip (DP: 10 steps, 1 sample) and maximum compute with \bon (Best-of-N: 10 steps, N samples). For \bon and when initializing samples in \ours, we use $N=8$ on the harder \textsc{Can Reverse} task and $N=4$ for other tasks. Appendix~\ref{ap:lf} also reports results for a \textit{learned} \fixed baseline, which directly optimizes a per-episode $(L, P)$ via a similar RL setup at the same $(\alpha, \beta)$ costs, rather than matching \ours's average allocation.

\begin{figure}[t!]
    \centering
    \includegraphics[width=\linewidth]{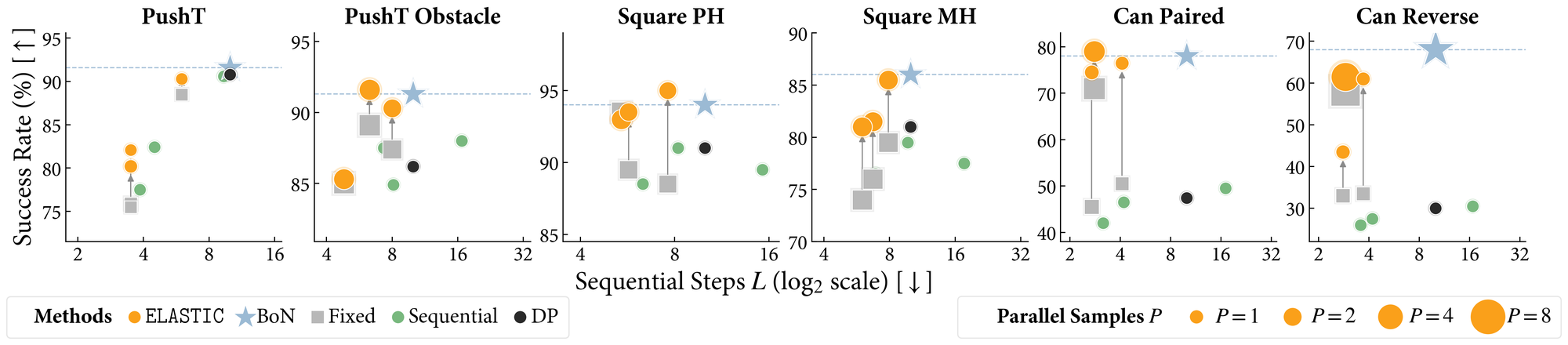}
    \caption{\textbf{Diffusion Policy Success Rate vs. Sequential Steps $L$ and Parallel Samples $P$.} \ours outperforms \fixed and \seq-only scaling at matched compute across 6 tasks and 3 cost settings (200 trials each), demonstrating the value of state-dependent allocation. Each marker is one method's operating point $(L, P)$; \seq and \fixed sweep three matched-budget points (one per cost setting). The gap between \fixed and \ours (gray arrow) is attributable to state-dependent allocation alone, since \fixed uses \ours's average $L, P$.}
    \label{fig:dp_results}
\end{figure}

\noindent \textbf{Results. }
Figure~\ref{fig:dp_results} shows the success rate of each method against the sequential steps $L$ and parallel samples $P$ it uses, with \seq and \fixed each shown at three budgets matched to \ours's three $(\alpha, \beta)$ settings: \seq spends each budget entirely on $L$, while \fixed splits it between $L$ and $P$ but holds the split fixed across states. \ours consistently outperforms \fixed allocation at matched compute (shown by the gray arrow), with the largest gains on tasks where $\dpolicy$ was trained on diverse, imperfect data, such as \textsc{Can Paired} and \textsc{Square MH}.

These results also demonstrate that knowing the right amount to scale each axis at test-time is not obvious. \textsc{PushT} scales well with additional sequential steps, but barely benefits from \bon search. Conversely, in \textsc{Can} and \textsc{PushT Obstacle} tasks, sequential scaling performance gains saturate quickly, while parallel samples can substantially close the gap to \bon success rates with far fewer sequential steps. In general, it is unclear a priori for some arbitrary set of costs $(\alpha, \beta)$ whether sequential or parallel compute is more beneficial.

\subsection{What Test-time Compute Schedule Does \ours's Meta-Policy Learn?}

Having established that \ours outperforms fixed allocation, we now qualitatively examine the allocation strategies the meta-policy discovers.

\begin{figure}[h!]
    \centering
    \includegraphics[width=\linewidth]{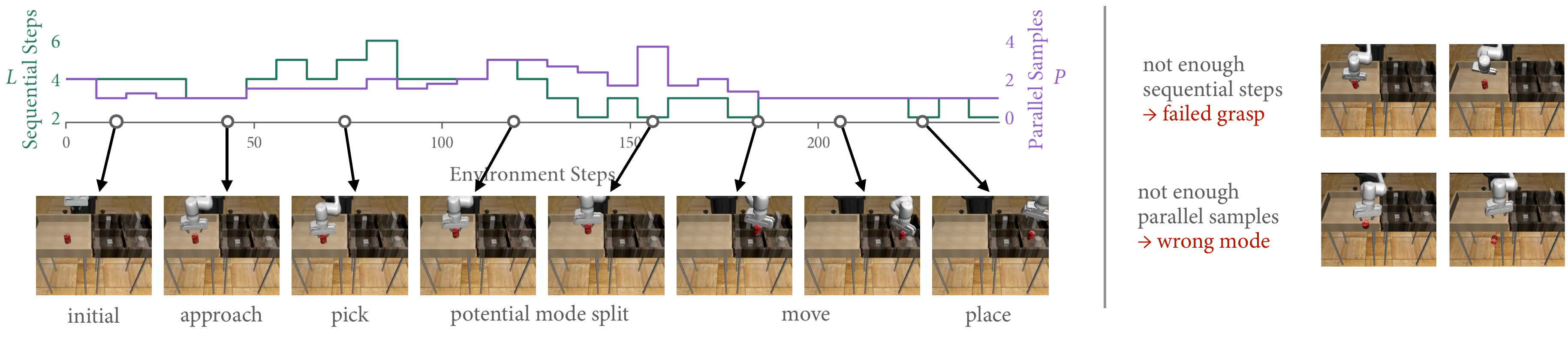} %
    \caption{\textbf{Allocation Timeline: \textsc{Can Paired}}. \ours allocates more sequential steps during the pick, to avoid a missed grasp (top right), and more parallel samples at states where action modes potentially split, to avoid selecting a wrong mode that throws the can off the table (bottom right).}
    \label{fig:can}
\end{figure}

\begin{figure}[H]
    \centering
    \includegraphics[width=\linewidth]{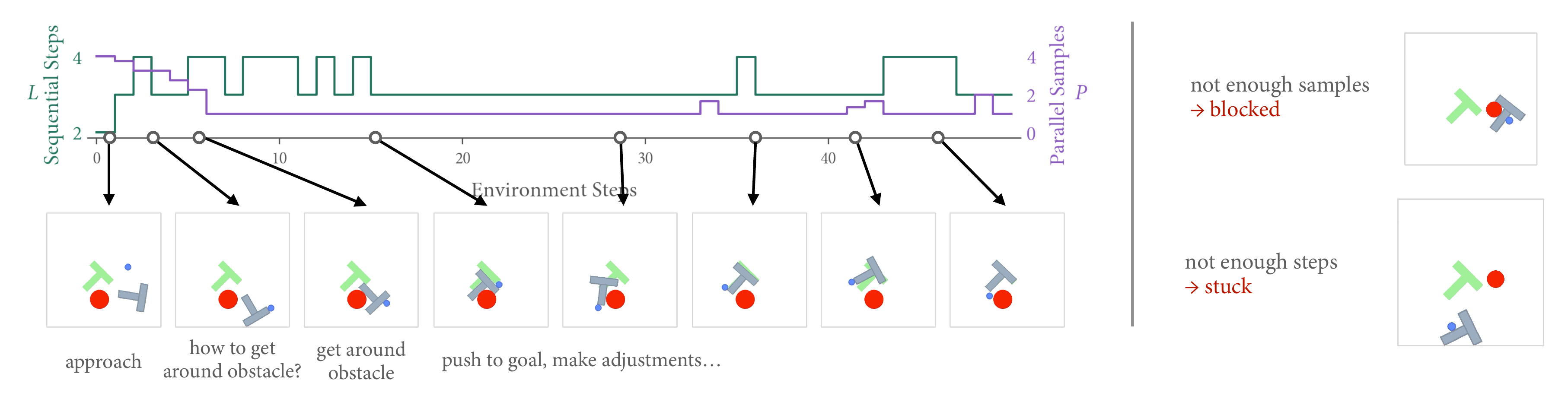}
    \caption{\textbf{Allocation Timeline: PushT Obstacle}. Using \ours, the meta-policy allocates more parallel samples at the very start, while the obstacle stands between the T and the goal. Sequential steps increase when the agent is actively pushing the T around the obstacle, as well as when making precise adjustments near the goal. Failure cases include getting blocked by the obstacle (not enough parallel samples) and getting stuck in a corner (not enough sequential steps).}
    \label{fig:pusht}
\end{figure}

\noindent \textbf{\ours autonomously discovers bottlenecks to task success.} \textsc{Can Paired}, \textsc{Can Reverse}, and \textsc{PushT Obstacle} show the clearest evidence of state-dependent allocation. For example, Figure~\ref{fig:can} shows how the meta-policy allocates compute along both axes across time in \textsc{Can Paired}. Sequential steps spike at the pickup phase, where clean denoising is required for reliable grasping; parallel candidates spike after, where the base policy's training data splits into multiple modes and $Q_\phi$ can discriminate between them. After the mode split resolves, both axes collapse: free motion toward the goal is unimodal and simple for the base policy. Since the mode split is concentrated at a small number of states, Figure~\ref{fig:dp_results} shows that \ours achieves success with substantially less total compute than full Best-of-N scaling at every step. Similar trends can be seen in \textsc{PushT Obstacle} (Figure~\ref{fig:pusht}), where \ours maximizes parallel sample usage to navigate around the obstacle.

\begin{figure}[H]
    \centering
    \includegraphics[width=\linewidth]{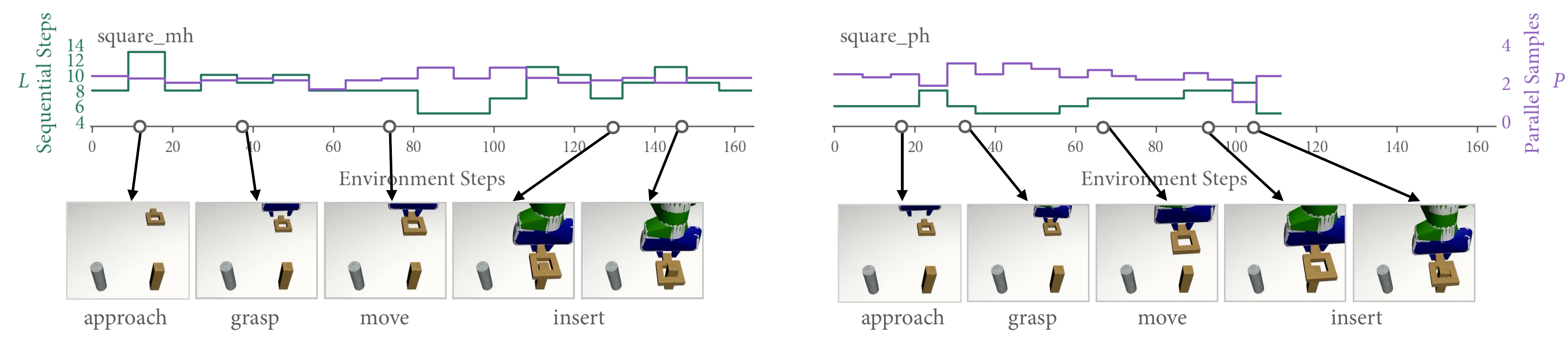}
    \caption{\textbf{Allocation Timeline: Square MH and PH}. Using \ours, the meta-policy allocates more parallel compute overall to Square MH (left) compared to Square PH (right). In particular, sequential compute varies a lot throughout the Square MH trajectory, peaking when first moving toward the square and while approaching the insertion. }
    \label{fig:square}
\end{figure}

\noindent \textbf{Compute allocation adapts to base policy performance, not just task geometry.} Despite sharing the same task environment, \textsc{Square PH} and \textsc{Square MH} yield different allocation patterns under the same cost coefficients and verifier: Figure~\ref{fig:square} shows that the MH checkpoint receives a larger number of sequential steps and parallel sample usage throughout, whereas the PH checkpoint does not, revealing where mixed-quality demonstrations impair the base policy. Notably, these states do not always correspond precisely to the grasp or insertion phases, suggesting that \ours learns to use test-time compute to compensate for regions where mixed data actually degrades task performance.

\subsection{\ours Yields \textit{Efficient} Test-time Scaling with Vision-Language-Action Models}

The GCPs we have studied condition on privileged, low-dimensional state vectors when generating actions. Now, we turn to state-of-the-art vision-language-action (VLA) model $\pi_{0.5}$~\cite{pmlr-v305-black25a}, which generates actions conditioned on visual observations, and study how \ours balances performance improvements against compute costs in both simulation and hardware. 

\begin{wrapfigure}[16]{R}{0.35\columnwidth}
    \vspace{-\intextsep}
    \centering
    \includegraphics[width=\linewidth]{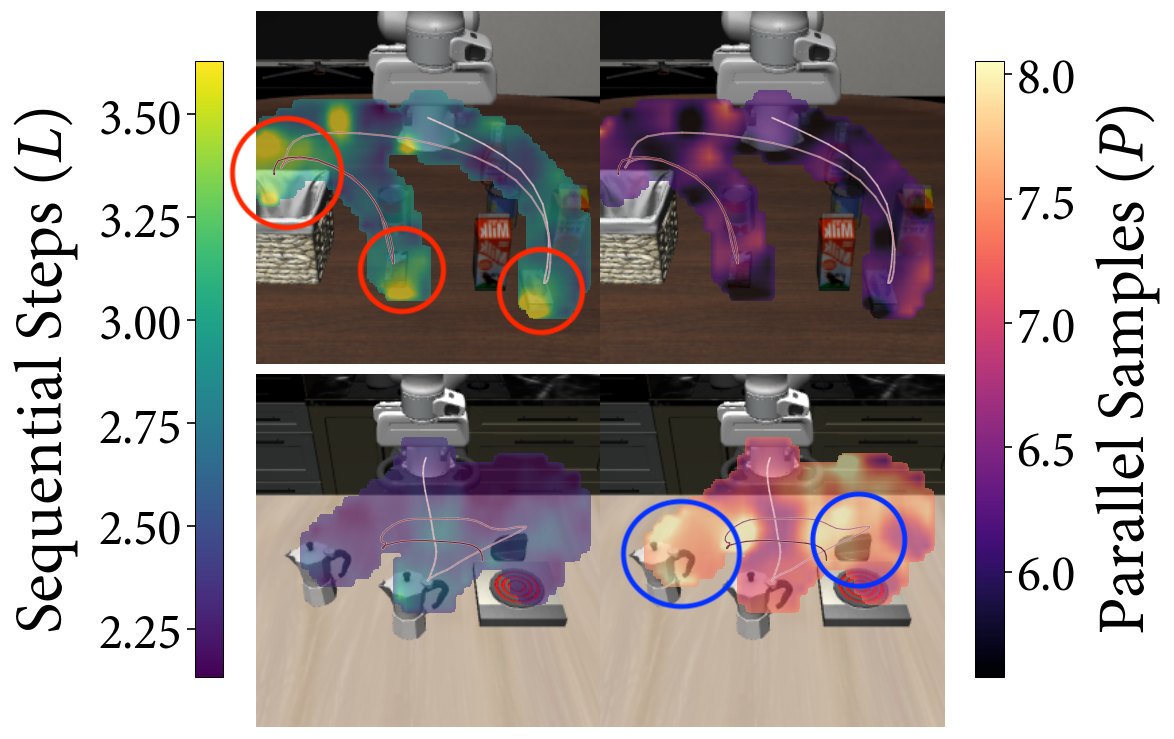}
    \caption{\textbf{LIBERO-10 Allocation Heatmaps.} \ours allocates compute differently across different tasks on a multi-task VLA. At pick and place states, the top task shows elevated sequential steps, whereas the bottom task instead uses more parallel samples.}
    \label{fig:libero-heatmap-mini} %
    \vspace{-\intextsep}
\end{wrapfigure}

\noindent \textbf{Setup.} For \textit{simulation}, we use the LIBERO-10 benchmark \cite{liu2023libero} and the corresponding task prompts and run the $\pi_{0.5}$-LIBERO checkpoint zero-shot over 20 trials. 
For \textit{hardware}, we use a Franka Emika robotic arm running the $\pi_{0.5}$-\textsc{DROID} checkpoint zero-shot. 
We design a real-world pick-and-place task where there are two mugs on a table (one purple, one blue) and the language prompt is \verb|place the purple mug in the basket|.
Over 20 trials, $\pi_{0.5}$ achieves 45\% success rate, as initial conditions in which the purple mug is further away from the camera tend to confuse the base policy $\pi_{0.5}$ about which mug to move toward. 
Moreover, the robot sometimes ends up in poor configurations during the place, unable to reach over the basket edges. 
For the verifier, we fine-tune V-GPS~\cite{nakamoto2024steering} on a mixture of success and failure demos of each task (details in Appendix~\ref{ap:v-gps}). Recall Section~\ref{sec:mdp} formalizes the meta-policy as having access to the true state $x$; here, we instead condition $\mpolicy$ on $\pi_{0.5}$'s observation embeddings, since privileged state is unavailable at deployment.
\begin{wrapfigure}[16]{R}{0.35\columnwidth}
    \vspace{-\intextsep}
    \centering
    \includegraphics[width=\linewidth]{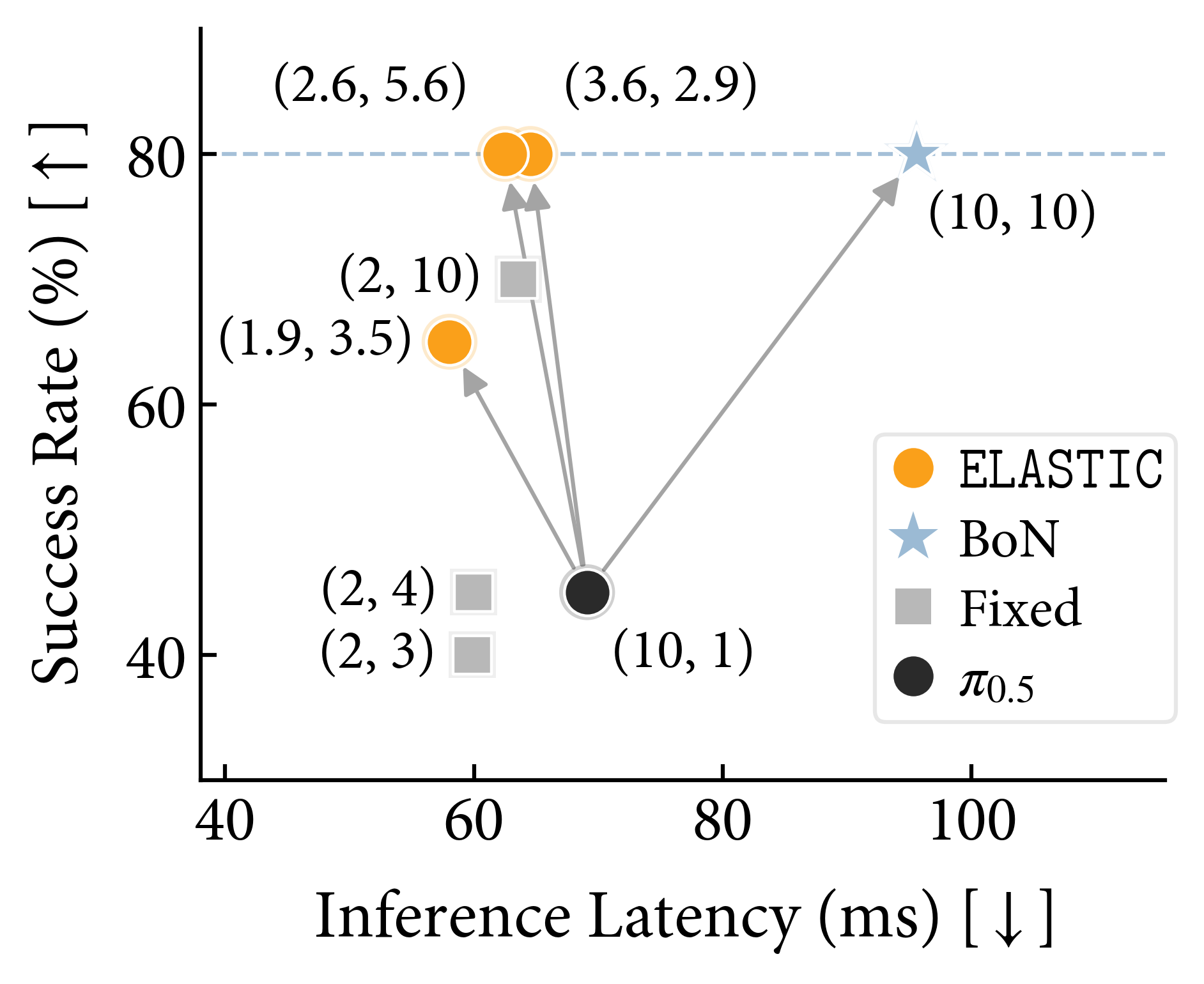}
    \caption{\textbf{Real World Success Rate vs Inference Latency.} \ours adaptively scales compute to match \bon performance with 34\% less inference latency. Average (Sequential Steps, Parallel Samples) per method are labeled.}
    \label{fig:realworld}
    \vspace{-1.0em}
\end{wrapfigure}
\noindent \textbf{Methods \& Metrics.} Similar to Sec. \ref{sec:dp} above, we measure success rate via binary task completion. For compute, we measure the wall-clock latency of each inference call (including $Q_\phi$ and $\mpolicy$ if applicable, specifics in Appendix~\ref{ap:timing}) measured on an RTX 4090 for simulation and RTX 6000 for hardware. In both simulation and hardware, we do 20 trials of each task per method. We compare \ours to the \textbf{Base} VLA, as well as to \bon and \fixed with the same verifier. We use the V-GPS default of $N=10$ samples for \bon and when initializing samples in \ours.

\noindent \textbf{Hyperparameters.} For \textit{simulation}, we set $\alpha=0.13$ and $\beta=0.1$ based on wall-clock latency scaling tests on a single RTX 4090. The optimization is conducted with 10 rollouts per task from the base policy for the offline dataset and 20 online rollouts per task.
For \textit{hardware}, we test our algorithm with 3 different compute costs: $(\alpha, \beta) = (0.13, 0.1), \ (0.2, 0.1), \ (0.13, 0.2)$.

\begin{figure}[H]
    \centering
    \includegraphics[width=0.7\linewidth]{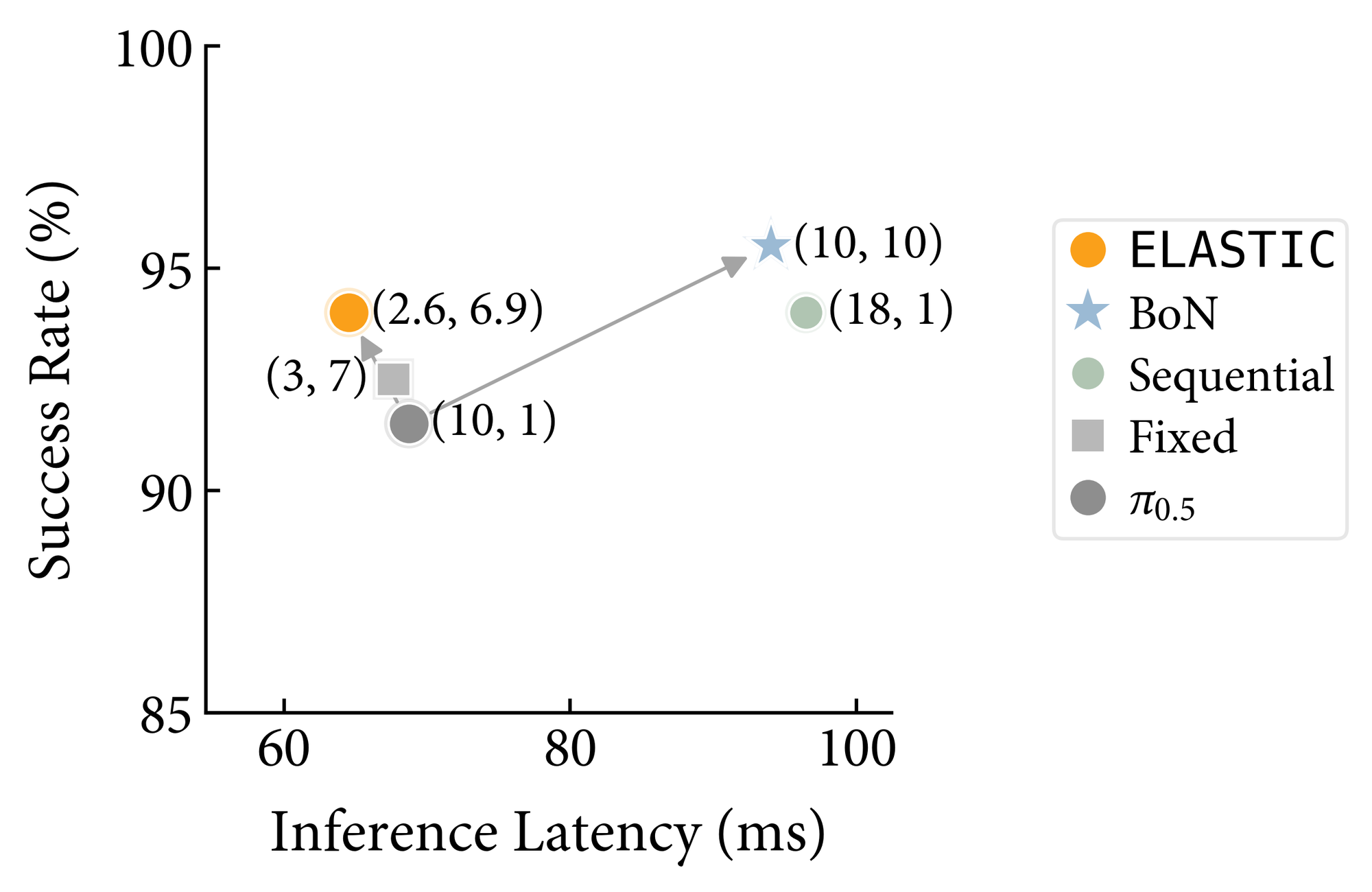}
    \includegraphics[width=\linewidth]{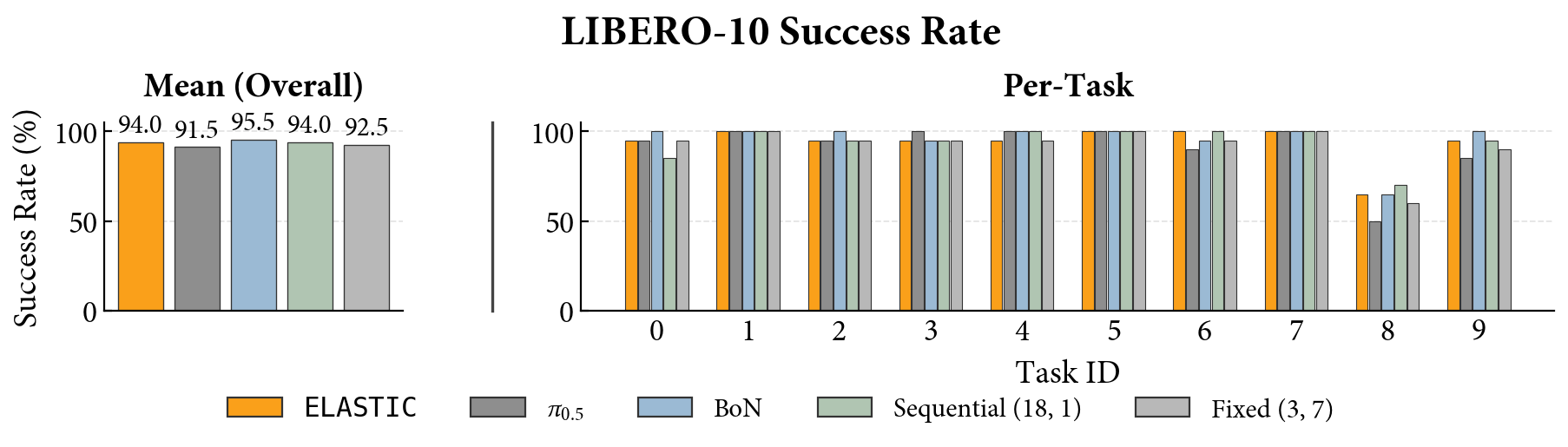}
    \includegraphics[width=\linewidth]{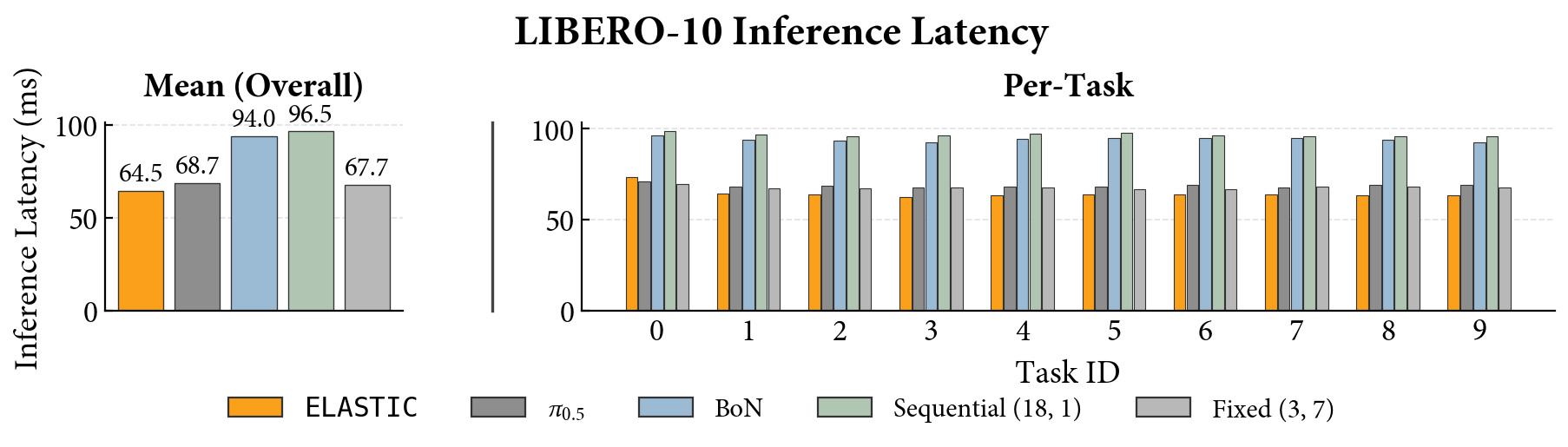}
    \caption{\textbf{LIBERO-10 Success Rates vs Inference Latency.} On LIBERO-10, base policy $\pi_{0.5}$ already achieves strong performance. Both scaling axes provide minor success rate gains at the cost of significantly longer inference latency: +36.8\% for \bon with V-GPS (10 Sequential Steps, 10 Parallel Samples) and +40.3\% for \seq (18 Sequential Steps, 1 Sample). \ours recovers performance gains equal to \seq with 6.1\% \textit{less} inference latency than the base $\pi_{0.5}$ policy, using (2.6 sequential steps, 6.9 parallel samples) on average.}
    \label{fig:libero_results}
\end{figure}

\noindent \textbf{Results: Simulation.} 
Figure~\ref{fig:libero_results} shows success rates and inference latency across all tasks and methods. Overall success rates are comparable across methods as $\pi_{0.5}$ already performs well. 
On tasks 8 and 9, where $\pi_{0.5}$ has a lower success rate, scaling either axis (parallel, \bon with V-GPS, or \seq) improves performance but significantly increases inference latency (+37--40\%).
\ours matches the higher performance of \seq with 6\% \textit{less} inference latency than the \textbf{Base} VLA, which defaults to a fixed denoising schedule with 10 sequential steps.

\begin{figure}[H]
    \centering
    \includegraphics[width=\linewidth]{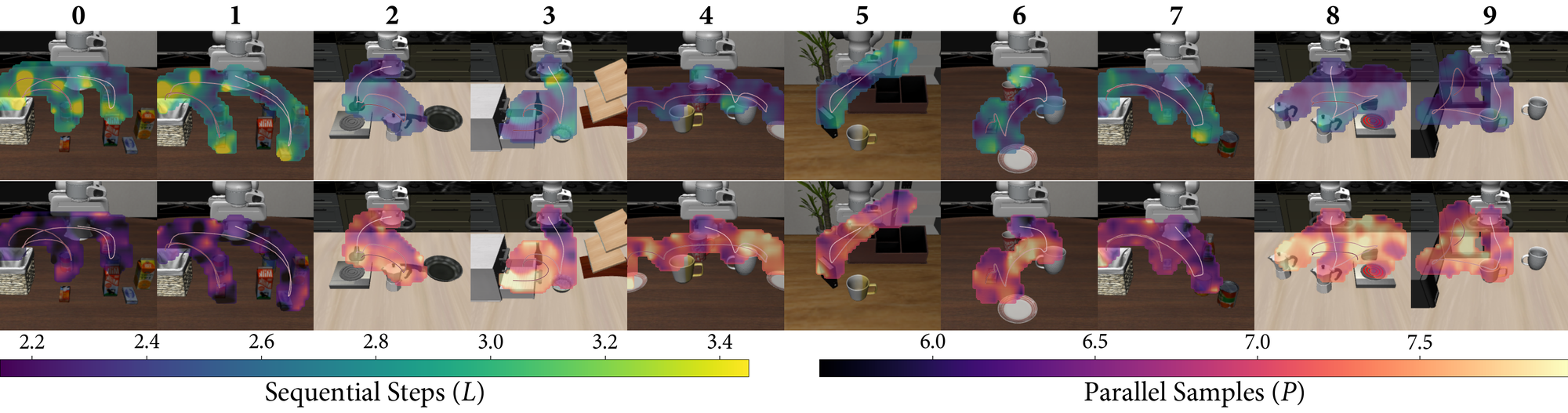}
    \caption{\textbf{LIBERO-10 Allocation Heatmap.} The heatmap shows how the \ours allocates sequential steps (top) and parallel samples (bottom) across states for $\pi_{0.5}$.}
    \label{fig:libero_latency_heatmap}
\end{figure}

Qualitatively, allocation patterns vary substantially across tasks and states, shown in Figures~\ref{fig:libero-heatmap-mini} and~\ref{fig:libero_latency_heatmap}. Tasks with high base policy success show low parallelism throughout, while tasks like 8 and 9, in which the base policy benefits from verifier usage, show higher parallelism near grasp and place states. Other tasks in which the \textbf{Base} VLA reaches 100\% success rate show reduced sequential steps throughout and elevated usage near the critical pick and place regions.

\begin{figure}[H]
    \centering
    \includegraphics[width=\linewidth]{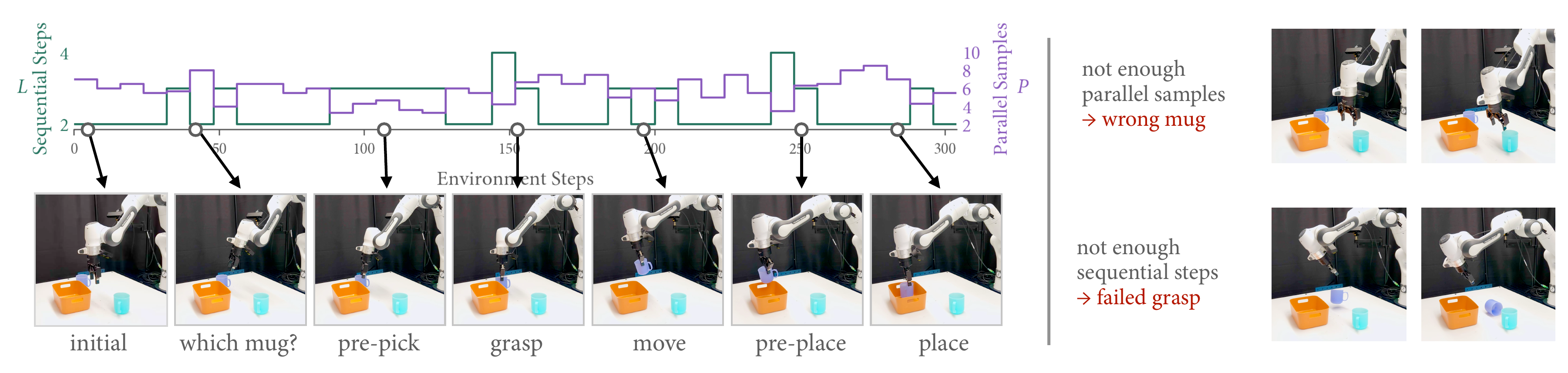}
    \caption{\textbf{Allocation Timeline: Real World Mug Task}. Using \ours, the meta-policy allocates more parallel samples when deciding which mug to pick near the start and during the placing action. Sequential steps vary less, with some peaks near grasping and prior to placing. Failure cases when using insufficient compute include selecting the wrong mug, failing to grasp, and failing to reach over the basket prior to placing.}
    \label{fig:real_timeline}
\end{figure}

\noindent \textbf{Results: Hardware.} Using \ours, the policy has less inference latency than the \textbf{Base} VLA while improving performance close to full scaling. Figure~\ref{fig:realworld} shows that \ours can match the improvements of full \bon scaling with V-GPS while using 34\% less inference latency. 
Achieving stronger state-conditioned performance requires additional interaction; here we used 60 offline demos and 30 online rollouts.
The allocation timeline in Figure~\ref{fig:real_timeline} shows that the number of sequential steps needed varies little throughout the task and is far fewer than the default of 10. Based on this result, we tested \fixed with (2 sequential steps, 10 samples) and found that it also achieves comparable performance to full \bon while using less latency than the \textbf{Base} VLA. In contrast, \fixed with (2 steps, 4 samples) performs worse than \ours at (1.9  steps, 3.5 samples) at comparable latency, suggesting that state conditioning becomes more valuable at smaller compute budgets.

\section{Limitations}
Although \ours consistently outperformed state-independent allocation and was able to improve base policy success while reducing inference latency, limitations remain.   
The primary limiting factor for the effectiveness of our approach is the learned verifier $Q_\phi$. 
A noisy verifier masks useful signal for distinguishing action quality, hurting meta-policy optimization and parallel scaling.
However, general-purpose reward models and verifiers are a rapidly advancing area, and our method demonstrates how to leverage these models efficiently, especially as they grow in computational cost \cite{liang2026robometer, ma2025vision, lee2026roborewardgeneralpurposevisionlanguagereward, tan2025robo, dong2026fastervalueguidedsamplingfast}.
Also, the meta-policy is currently trained per task, which can be costly for multi-task policies like VLAs. Since it conditions on the base policy's internal features and action samples, a promising direction is to test whether these signals can generalize to unseen environments.

\newpage 
\section{Appendix}

\subsection{\ours Algorithm}
\begin{algorithm}
\caption{Inference at every environment step $t$}
\begin{algorithmic}[1]
\State Observation $o_t$, base policy $\dpolicy$, meta-policy $\mpolicy$, reward model $Q_\phi$, $N$ samples

\State Initialize $\ctrl^{(i)} \sim \mathcal{N}(0, I)$ for $i = 1, \dots, N$
\State Initialize $\ntime^{(i)} \leftarrow 1$ for $i=1,\dots,N$
\State Initialize $m^{(i)} \leftarrow 1$ for $i=1,\dots,N$
\\
\While{$\exists\, i$ such that $\ntime^{(i)} > 0$ and $m^{(i)} = 1$}
    \State $s=(o_t, u^{1:N}, \ntime^{1:N}, m^{1:N})$
    \State $\stepsize^{1:N} \leftarrow \mpolicy(s)$ \Comment{Joint denoising decision over all $N$ samples}
    \\
    \For{$i = 1, \dots, N$} \Comment{Vectorized over $N$}
        \If{$m^{(i)} = 0$ or $\ntime^{(i)} = 0$}
            \State \textbf{continue} \Comment{Sample already stopped}
        \ElsIf{$\stepsize^{(i)} = 0$}
            \State $m^{(i)} \leftarrow 0$ \Comment{Permanently stop sample}
        \Else
            \State $\ctrl^{(i)} \leftarrow \dpolicy(o_t, \ctrl^{(i)}, \ntime^{(i)}, \stepsize^{(i)})$ \Comment{Denoising step}
            \State $\ntime^{(i)} \leftarrow \max(\ntime^{(i)} - \stepsize^{(i)}, 0)$
        \EndIf
    \EndFor
    \\
\EndWhile
\\
\State $\mathcal{F} \leftarrow \{i : \ntime^{(i)} = 0\}$ \Comment{Filter to fully denoised samples only}
\State $\hat \ctrl_t \leftarrow \arg\max_{i \in \mathcal{F}}\, Q_\phi(o_t, \ctrl^{(i)})$ \Comment{Select best with verifier}
\State Execute $\hat \ctrl_t$ in environment, observe $o_{t+1}$
\end{algorithmic}
\label{alg:inference}
\end{algorithm}

\begin{figure}[H]
    \centering
    \includegraphics[width=\linewidth]{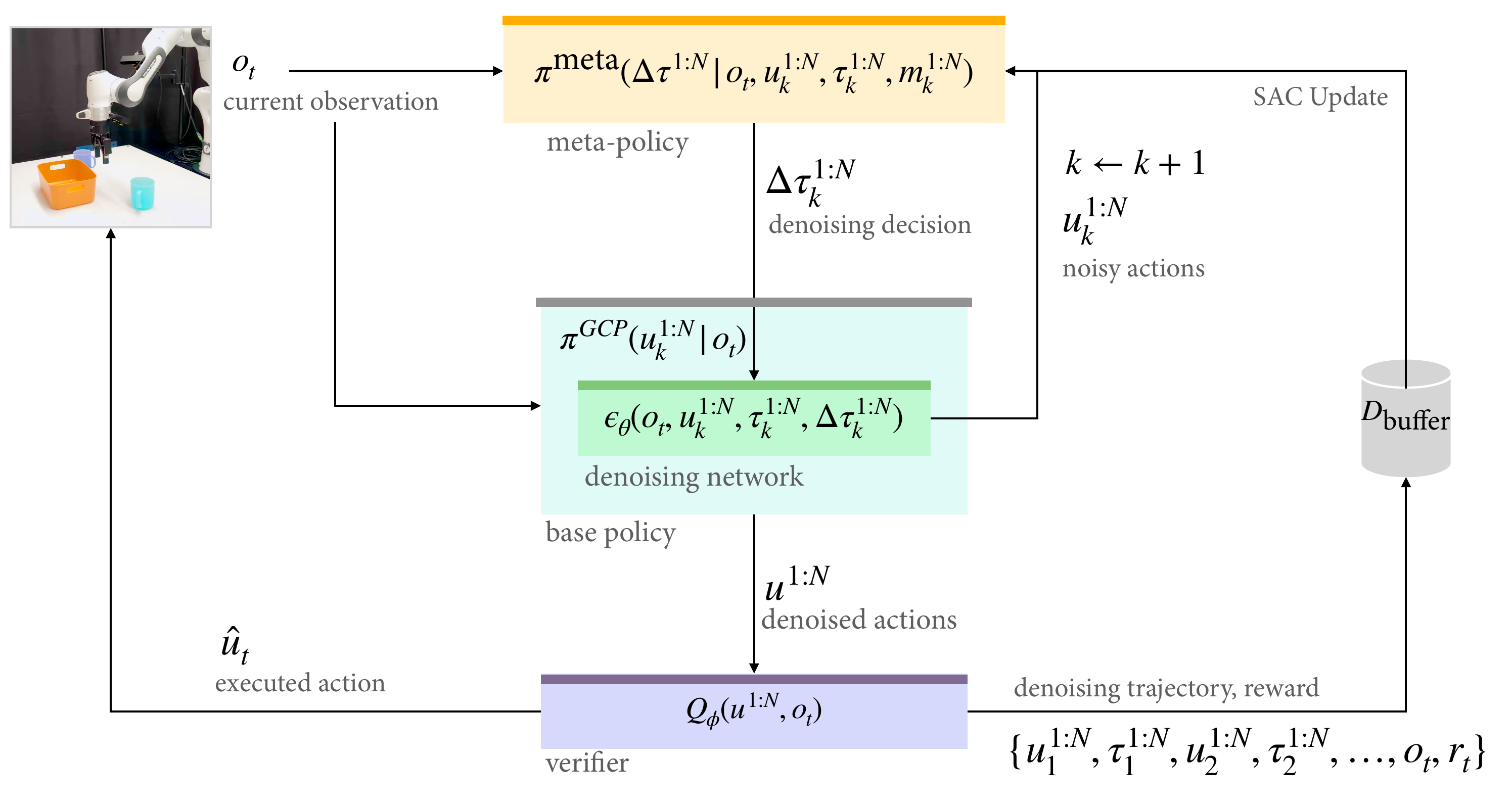}
    \caption{\textbf{\ours Pipeline.}}
    \label{fig:inference}
\end{figure}

\subsection{Additional Results}

\begin{figure}[H]
    \centering
    \includegraphics[width=\linewidth]{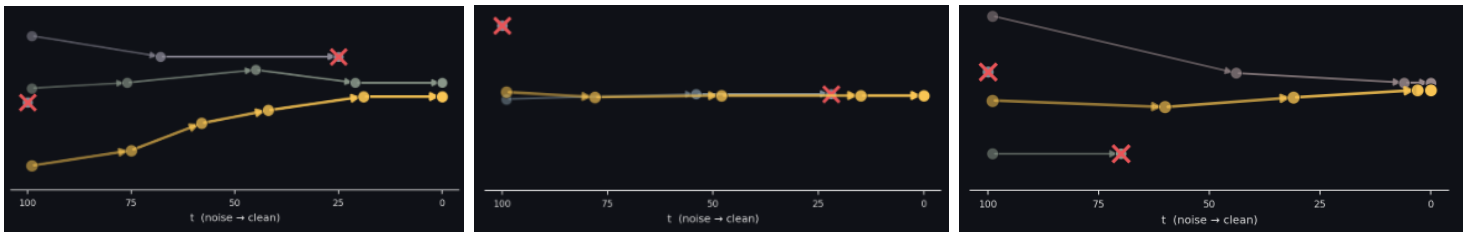}
    \caption{\textbf{Denoising Timeline.} Visualization of how action samples evolve through the denoising process under the meta-policy. Each plot shows the Euclidean distance between samples vs. denoising time from different states in a \textsc{Can Paired} trajectory. Stopping behaviors emerge from the single scalar reward in Eq.~\ref{eq:reward} without hand-designed stopping rules or specially trained noise-aware verifiers.}
    \label{fig:coordination}
\end{figure}

\subsubsection{Baselines}
\label{ap:lf}
Since \ours uses varying $(L, P)$ per state, \fixed matches the mean, $\bar L, \bar P$, by randomly sampling between $\{\lfloor \bar L\rfloor, \lceil \bar L \rceil\}$ steps and $\{\lfloor \bar P\rfloor, \lceil \bar P \rceil\}$ samples at each environment state, preserving the average while remaining state-independent.

\seq converts the learned policy’s average compute budget into an equivalent single-sample denoising budget $L=(\alpha \bar L + \beta (\bar P-1))/ \alpha, P=1$, and similarly applies floor/ceil rounding so that the realized compute is matched.

\textit{Learned} \fixed uses the same architecture and training setup as \ours but optimizes the $(L,P)$ action space. This controls for the possibility that \fixed underperforms due to a suboptimal allocation along each axis rather than its lack of state-dependence. However, note that this results in different amounts of compute used along each axis compared to \ours. 

\begin{figure}
    \centering
    \includegraphics[width=\linewidth]{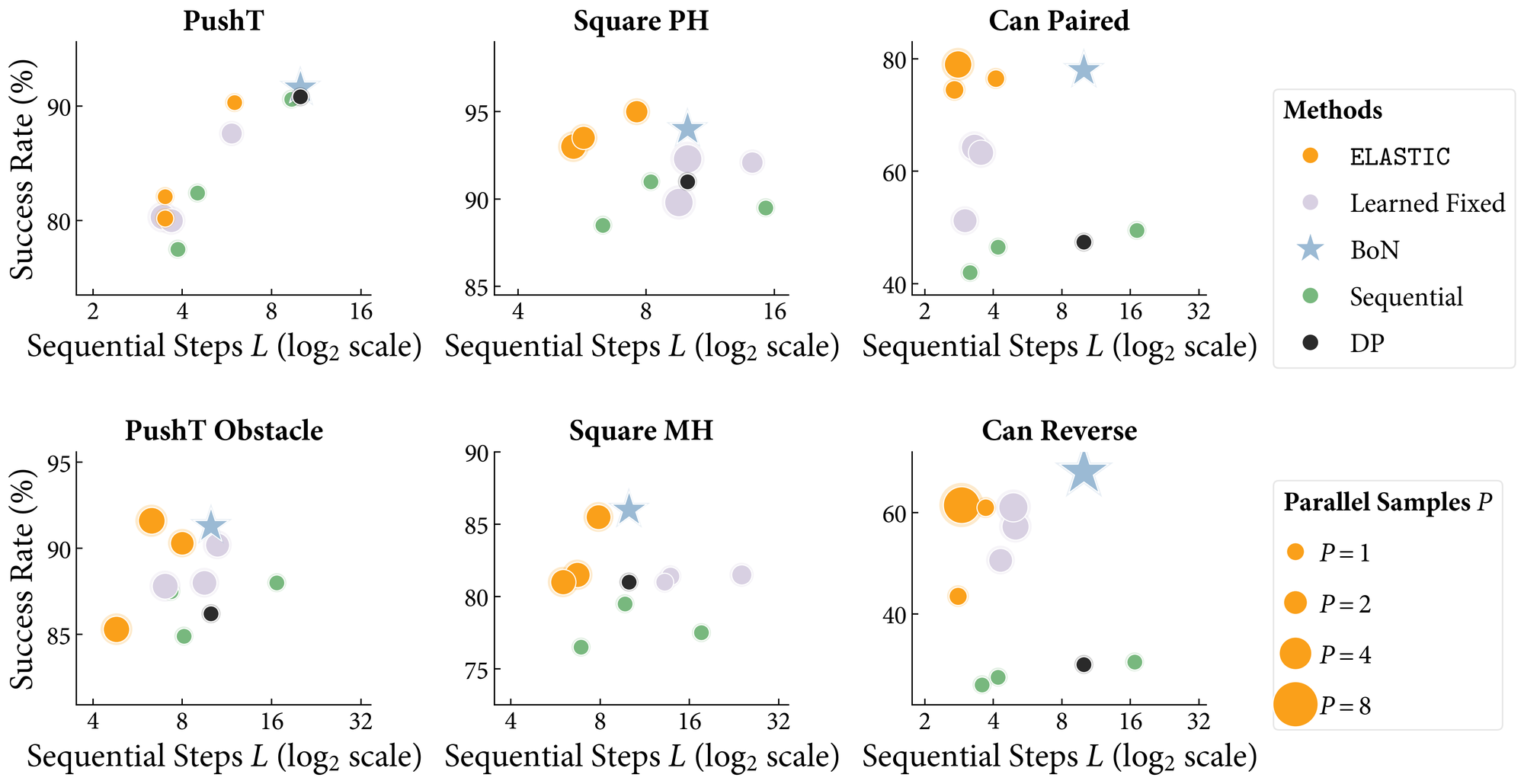}
    \caption{\textbf{Diffusion Policy Success Rate vs. Sequential Steps $L$ and Parallel Samples $P$.} We plot an additional \textit{Learned} \fixed baseline, which learns a constant allocation of $L$ sequential steps and $P$ parallel samples per action generation.}
    \label{fig:learned_fixed}
\end{figure}

\subsection{Meta-Policy Training Details}
\label{ap:subsample}

To improve sample efficiency, we augment the online replay buffer with 
counterfactual compute allocations at each visited state. Every $C=4$ 
environment steps, we run a second pass of $\pi^{\texttt{base}}$ with all 
$N$ candidates active at a fixed stride $\stepsize \sim 
\text{Uniform}(0.5\overline{\stepsize},\, 1.5\overline{\stepsize})$, where 
$\overline{\stepsize}$ is a running average of $\mpolicy$'s decisions.

All $N$ fully-denoised candidates are scored with $Q_\phi$. We then generate 
synthetic terminal transitions for subsample sizes $k \in \{1, 2, 4, \ldots, 
N\}$ by randomly selecting $k$ candidates. For each subsample, the terminal 
reward is recomputed via Eq.~\ref{eq:reward}: task quality is the maximum 
verifier score among the $k$ retained candidates, the sequential cost is fixed by $\stepsize$, and the parallel cost reflects $k$ rather than $N$.  All tuples are added to the shared replay buffer, providing coverage across the full range of parallel compute budgets and similar sequential compute budgets without additional environment interaction.

\subsection{Scaling Performance and Inference Time Measurement}
\label{ap:timing}
We measure inference latency as the wall-clock round-trip time of an \verb|infer()| request in $\pi_{0.5}$ over the same setup used during deployment. For each setting, we start a fresh policy server and run untimed warmup inference calls to exclude compilation effects. The measured inference latency is the client-side round-trip time, including client request overhead, meta-policy inference, base policy inference, and verifier inference.

For reference, we measure the inference costs of sequential scaling and parallel scaling on our real hardware setup, which uses an RTX 6000. We sweep $P$ from 1 to 10 with $L$ fixed at 1, 5, and 10 and average over 10 timed inference calls. We see that increasing parallel samples $P$ also increases the inference latency, and that the marginal cost increases with more sequential steps.

\begin{figure}[H]
    \centering
    \includegraphics[width=0.8\linewidth]{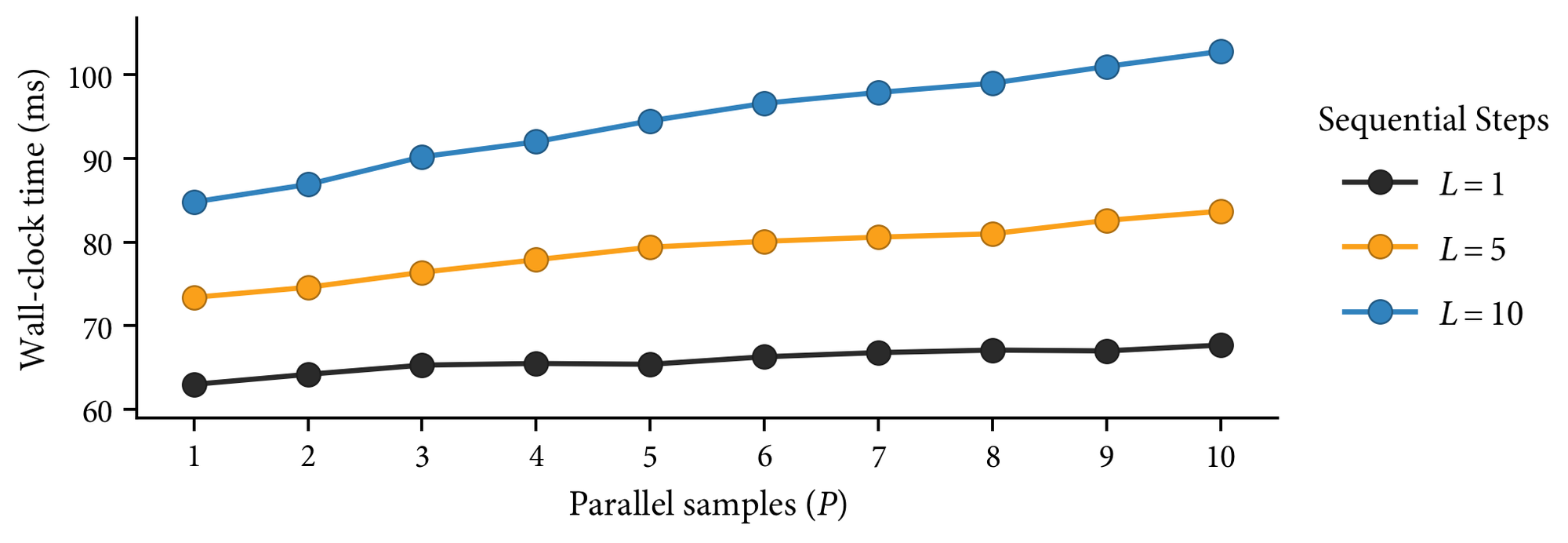}
    \caption{Inference Latency vs Sequential Steps and Parallel Samples}
    \label{fig:real_compute_costs}
\end{figure}

\subsection{Verifier Training}
Because $Q_\phi$ depends on environment state, reward targets can be high variance across different states. This can degrade sample efficiency, because the optimization can conflate good samples with good state. To address this, we find that computing the advantage $A = Q-V$ can help, either online using actions from the second $\dpolicy$ pass described in \ref{ap:subsample} as the baseline or offline by fitting $V$.

\subsubsection{Diffusion Policy Verifier Training}
\label{ap:dp-verifier}
We collect a dataset $\mathcal{D}_{\mathrm{ref}}$ using 200 rollouts of the base diffusion policy checkpoint at the default of 10 DDIM steps, which serves as the reference continuation policy. We train a simple two-layer MLP network with hidden width 128, followed by separate value and advantage heads, since we use advantage for the meta-policy's rewards. The predicted value $V_\theta(s)$ and advantage $A_\theta(s,a)$ are summed to $Q_\theta(s,a)=V_\theta(s)+A_\theta(s,a)$ and regressed against the empirical return-to-go training labels $\hat G$. We anchor the advantage by adding a loss term that encourages the mean advantage to be zero: $\mathcal{L}
=
\mathbb{E}_{(s,a,\hat{G})\sim\mathcal{D}_{\mathrm{ref}}}
\left(Q_\theta(s,a)-\hat{G}\right)^2
+
w_{\text{anchor}}\left(\mathbb{E}_{(s,a)\sim\mathcal{D}_{\mathrm{ref}}}
[A_\theta(s,a)]\right)^2
$

For certain tasks, the meta-policy is prone to exploiting the verifier by producing actions that the verifier scores highly but results in poor actual task returns. Using LayerNorms and adding a conservative Q-value regularizer to penalize high Q-values on randomly sampled action chunks helps address this: $\mathcal{L}_{\text{cons}}=w_{\text{cql}} \left(
\tau\log\sum_{a'\in\mathcal{A}(s)} e^{Q_\theta(s,a')/\tau}
-
Q_\theta(s,a)
\right)$, where $\mathcal{A}(s)$ contains actions from the dataset and randomly sampled action chunks.

\begin{table}[H]
    \centering
    \begin{tabular}{lc}
    \toprule
    \textbf{Hyperparameter} & Value \\ 
    \midrule
    Learning Rate & $1\times 10^{-4}$\\
    Weight Decay & $1\times 10^{-4}$ \\
    Batch Size & 256 \\
    Epochs & 200 \\
    $w_{\text{anchor}}$ & 0.1 \\
    $w_{\text{cql}}$ & 0.1 \\
    $\tau$ & 1 \\
    \bottomrule
    \end{tabular}
    \caption{Hyperparameters for the diffusion policy task verifiers.}
    \label{tab:verifier_hparam}
\end{table}

\subsubsection{V-GPS Fine-tuning}
\label{ap:v-gps}
We fine-tune V-GPS using the default Conservative Q-Learning (CQL) training configuration for LIBERO simulation tasks and Calibrated Q-Learning (Cal-QL) training configuration for the real task with default hyperparameters. We find that without Cal-QL on the real task, the verifier cannot distinguish between actions that lead to failure modes (pick up blue mug instead of the purple mug). We use the exterior view image from $\pi_{0.5}$ as the observation to V-GPS. We collected 30 rollouts per task on LIBERO and 120 rollouts for the real mug task.

\subsection{Meta-Policy Hyperparameters}
\begin{table}[H]
    \centering
    \begin{tabular}{lccc}
    \toprule
    \textbf{Hyperparameter} & Diffusion Policy & $\pi_{0.5}$-LIBERO & $\pi_{0.5}$-DROID \\ 
    \midrule
    Actor Learning Rate & $1\times 10^{-4}$ & $1\times10^{-4}$ & $3\times10^{-4}$ \\
    Critic Learning Rate & $3\times10^{-4}$ & $3\times10^{-4}$ & $3\times10^{-4}$ \\
    Temperature Learning Rate & $3\times10^{-4}$ & $3\times10^{-4}$ & $3\times10^{-4}$ \\
    Discount & 1 & 1 & 1 \\
    UTD Ratio & 30 & 8 & 8 \\ 
    Batch Size & 512 & 256 & 256 \\
    Offline Pretraining Steps & 5000 & 5000 & 5000 \\
    Subsample Counterfactuals Every & 4 env steps & 4 env steps & 4 env steps \\ 
    \bottomrule
    \end{tabular}
    \caption{Hyperparameters for the meta-policy across 3 different base GCPs.}
    \label{tab:meta_hparam}
\end{table}

\end{document}